\documentclass{article}

    \PassOptionsToPackage{numbers, compress}{natbib}
\usepackage{multirow}
\usepackage{makecell}
\usepackage[final]{neurips_2024}

\usepackage[utf8]{inputenc} % allow utf-8 input
\usepackage[T1]{fontenc}    % use 8-bit T1 fonts
\usepackage{hyperref}       % hyperlinks
\usepackage{colortbl}  
\usepackage{xcolor}
\usepackage{url}            % simple URL typesetting
\usepackage{booktabs}       % professional-quality tables
\usepackage{amsfonts}       % blackboard math symbols
\usepackage{nicefrac}       % compact symbols for 1/2, etc.
\usepackage{microtype}      % microtypography
\usepackage{xcolor}         % colors
\usepackage{multicol}
\usepackage{graphicx}
\usepackage{bbm}
\usepackage{wrapfig}
\usepackage{amsmath}
\usepackage{threeparttable}

\title{Mixture of In-Context Experts \\Enhance LLMs' Long Context Awareness}

\author{%
  Hongzhan Lin$^{1}$\thanks{Equal contribution. 
  Hongzhan Lin and Ang Lv proposed the idea of MoICE. 
  Hongzhan Lin and Yuhan Chen designed the MoICE router architecture and implemented efficient code. 
  Experiments were conducted by Hongzhan Lin, while Ang Lv led the writing.
  Code is available at \url{https://github.com/p1nksnow/MoICE}.
  } 
  \quad Ang Lv$^{1}$\footnotemark[1] \quad Yuhan Chen$^{2}$\footnotemark[1]  \\
  \textbf{Chen Zhu}$^{3}$\quad \textbf{Yang Song}$^{4\dagger}$\quad \textbf{Hengshu Zhu}$^{3}$ \quad \textbf{Rui Yan}$^{1}$\thanks{Corresponding authors: Rui Yan (\url{ruiyan@ruc.edu.cn}) and Yang Song (\url{songyang@kanzhun.com})} \\
  $^{1}$ Gaoling School of Artificial Intelligence, Renmin University of China\\
  $^{2}$ XiaoMi AI Lab
  $^{3}$ Career Science Lab, BOSS Zhipin 
  $^{4}$ NLP Center, BOSS Zhipin \\
  \texttt{\{linhongzhan, anglv, ruiyan\}@ruc.edu.cn} \\
  \texttt{\{chenyuhan5\}@xiaomi.com} \\
}

\begin{document}

\maketitle

\begin{abstract}
Many studies have revealed that large language models (LLMs) exhibit uneven awareness of different contextual positions.
Their limited context awareness can lead to overlooking critical information and subsequent task failures. 
While several approaches have been proposed to enhance LLMs' context awareness, achieving both effectiveness and efficiency remains challenging.
In this paper, for LLMs utilizing RoPE as position embeddings, we introduce a novel method called ``Mixture of In-Context Experts'' (MoICE) to address this challenge. 
MoICE comprises two key components: a router integrated into each attention head within LLMs and a lightweight router-only training optimization strategy:
(1) MoICE views each RoPE angle as an `in-context' expert, demonstrated to be capable of directing the attention of a head to specific contextual positions.
Consequently, each attention head flexibly processes tokens using multiple RoPE angles dynamically selected by the router to attend to the needed positions.
This approach mitigates the risk of overlooking essential contextual information.
(2) The router-only training strategy entails freezing LLM parameters and exclusively updating routers for only a few steps.
When applied to open-source LLMs including Llama, Mistral and Qwen, MoICE surpasses prior methods across multiple tasks on long context understanding and generation, all while maintaining commendable inference efficiency.
Moreover, we also demonstrate the effectiveness of MoICE in pre-training a language model from scratch.
% We promise to open our code\footnote{\url{https://anonymous.4open.science/r/MoICE-13D6}}.
\end{abstract}

% A head then aggregates all attention scores to yield a comprehensive final attention result. 

\section{Introduction}
\label{sec:intro}

Although large language models (LLMs) have demonstrated impressive capabilities across diverse NLP tasks, several studies~\cite{liu2023lost,chen2023fortify,lv2023fallingmiddleintelligencetrapanalysis} have pointed out that the contextual awareness of LLMs is not as powerful as widely believed, constraining their application in tasks demanding extensive contextual awareness, such as in-context learning~\cite{lu-etal-2022-fantastically,zhang-etal-2024-batch}, coherent long text generation~\cite{zhang2024found} and Retrieval-Augmented Generation (RAG, ~\cite{huang2024survey,chen2024benchmarking,cheng2024lift}) tasks necessitating in-context retrieval~\cite{chen2023fortify}.
Liu et al.~\cite{liu2023lost} identified a common issue termed the ``lost-in-middle'' phenomenon, indicating that LLMs often exhibit a weaker awareness of information situated in the middle of the long context compared to the beginning or end. 
Chen et al.~\cite{chen2023fortify} highlighted challenges arising from a mathematical property of RoPE~\cite{su2024roformer}, a wide-used positional embedding in LLMs, which impedes attention to specific positions within the long context.
Consequently, if critical information coincides with such positions, task performance suffers.

\begin{figure}[t]
\vspace{-5mm}
    \centering
    \includegraphics[width=0.92\linewidth]{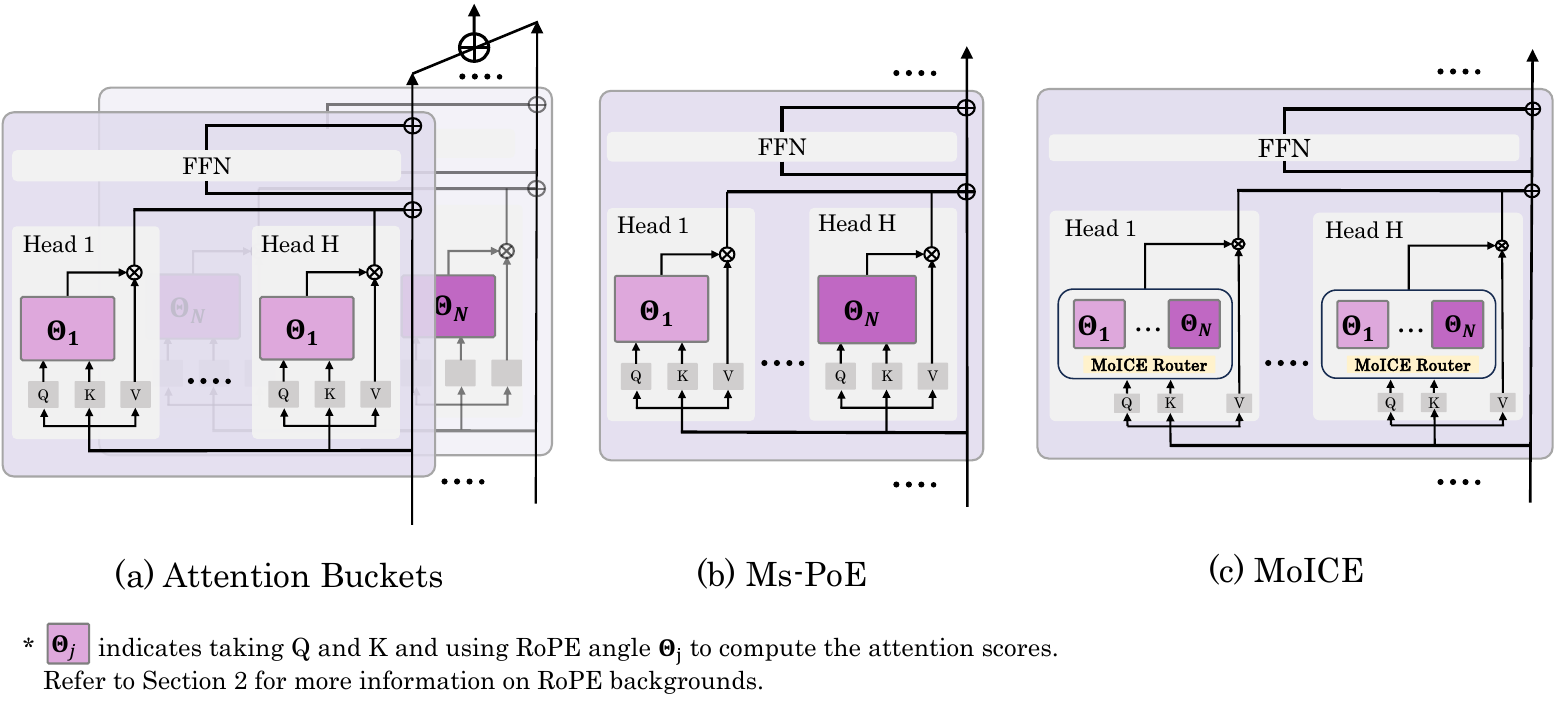}
    \caption{Some methods developed to enhance LLMs' context awareness. 
    (a) Attention Buckets~\cite{chen2023fortify} selects $N$ different RoPEs and conducts $N$ parallel inferences for each input.
    The outputs are then aggregated in the final layer.
    (b) Ms-PoE~\cite{zhang2024found} employs a unique RoPE angle for each attention head. 
    However, it needs an additional forward pass for RoPE angle assignment.
    (c) MoICE integrates a router within each attention head. 
    This novel plug-in selects several of the most suitable RoPE angles for each token.
    The selected RoPE angles collectively contribute to computing the attention scores. 
    MoICE demonstrates superior memory efficiency and performance.
    }
    \label{fig:intro}
    \vspace{-3mm}
\end{figure}

Many works~\cite{junqing2023never,zhang2024found,chen2023fortify,zhang2023tell} have attempted to enhance the long-context awareness of LLMs. 
Central to these efforts is the enhancement of attention heads which serve as the linchpin for contextual awareness, given that FFNs in language models do not introduce token interaction. 
Chen et al.~\cite{chen2023fortify} proposed an inference algorithm named \textit{Attention Buckets} (AB), which enhanced the context awareness of LLMs by executing $N$ inference instances, each with a distinct RoPE angle, and aggregated the outputs at the final layer.  
Zhang et al.~\cite{zhang2024found} observed the varying awareness of attention heads to contextual positions. 
They proposed an inference algorithm named \textit{Ms-PoE}. 
Ms-PoE enhances the utility of position-aware heads by re-scaling the positional embedding indices, equivalent to assigning each head a unique RoPE angle. Figure~\ref{fig:intro} illustrates these approaches. 
However, these approaches each come with their own drawbacks: AB conducts excessive redundant FFNs calculations, leading to high memory consumption. 
In Ms-PoE, determining a distinct re-scale factor for every attention head needs an additional forward pass. 
Meanwhile, each attention head still depends on a single re-scaled static RoPE.
As highlighted by AB~\cite{chen2023fortify}, this leads to limited awareness of certain contextual positions, thereby constraining its potential.
Moreover, a significant drawback of both AB and Ms-PoE lies in their static assignment of the RoPE angle for each attention head throughout the generation. However, as the generation progresses, the positions of crucial tokens shift, necessitating corresponding adjustments in the required RoPE angles for each head.

In this study, we present \textit{Mixture of In-Context Experts} (MoICE), a novel plug-in of LLMs for enhancing context awareness.
Specifically, We conceptualize a unique RoPE angle as an ``in-context expert,'' as it can allocate a head's more attention to certain contextual positions~\cite{chen2023fortify}. 
We integrate a router within each attention head, which discerns the potentially important tokens for the head and dynamically selects $K$ RoPE angles that provide comprehensive awareness of these tokens for attention computation.
Through the re-computation of only a few query-key dot products, attention patterns computed with selected RoPE angles are aggregated to produce the final attention pattern.
This approach yields two primary advantages: 
(1) It eliminates unnecessary computational overhead in AB, enhancing efficiency.
(2) The dynamic expert selection of each head for arbitrary tokens introduces flexibility not attained in previous studies. This minimizes the risk of the initial RoPE angle assigned to a head failing to work due to crucial token positions shifting during generation.

Consequently, MoICE not only surpasses AB's effectiveness but also achieves commendable efficiency.
We name our approach as ``Mixture of In-Context Experts'' (MoICE) due to the aggregation of attention patterns calculated with different RoPE angles resembling the concept of ``Mixture of Experts'' (MoE,~\cite{shazeer2017}).
When applying MoICE to open-source LLMs, we freeze LLMs' parameters and conduct lightweight training only on the MoICE routers.
%这个细节放后面讲？只要说推理和训练快就行？
%By training for only 8 minutes on four A800-80G GPUs
With only a few quick updates, MoICE surpasses many competitive baselines in tasks involving long-context generation and understanding.

In summary, our main contribution is the introduction of MoICE, a novel plug-in for enhancing LLMs' context awareness. 
It achieves head-and token-specific dynamic multiple RoPE angles assignment, outperforms previous methods across various tasks, and maintains commendable inference efficiency.

\section{Background}
We introduce some background of \textit{Mixture of In-Context Experts}, including (1) the rotary position embeddings commonly used by mainstream LLMs, (2) the primary problem addressed in this paper: the limited context awareness of LLMs, (3) an explanation of the underlying reasons for this limitation, and (4) the Mixture of Expert techniques employed to mitigate the limitation.

\paragraph{Position embedding}
Positional embedding is crucial for Transformer~\cite{vaswani} to perceive sequence order and compensate for the position-agnostic nature of the attention mechanism. 
In this paper, we mainly focus on LLMs using Rotary Position Embedding (RoPE,~\cite{su2024roformer}) which is the prevalent position embedding in current LLMs.
We discuss other position embeddings in Appendix~\ref{apx:other-pos}.

In a Transformer layer with $H$ attention heads employing RoPE, where $d$ represents the hidden state dimension of each attention head, let $\mathbf{q}^{h}_n$ and $\mathbf{k}^{h}_m$ denote the query vector at position $n$ and key vector at position $m$ in the $h$-th head. 
To encode position information, RoPE initially applies a rotary matrix to the query and key vectors:
\begin{equation}
    \mathbf{\hat{q}}_n^{h}=\mathbf{R}_{\Theta_j,n}\cdot \mathbf{q}_n \in \mathbb{R}^{d}, \quad \mathbf{\hat{k}}_m^{h}=\mathbf{R}_{\Theta_j,m}\cdot \mathbf{k}_m \in \mathbb{R}^{d},
\end{equation}
\begin{equation}
\resizebox{0.98\linewidth}{!}{
$\mathbf{R}_{\Theta_j, n}=\left[\begin{array}{cccc}
\mathbf{r}_{\theta_{j,0}, n} & O & \cdots & O \\
O & \mathbf{r}_{\theta_{j,1}, n} & \cdots & O \\
\vdots & \vdots  & \ddots & \vdots   \\
O & O & \cdots & \mathbf{r}_{\theta_{j,d/2-1}, n} \\
\end{array}\right], \text{\ where\ \ } \mathbf{r}_{\theta_{j,i}, n}=\left[\begin{array}{cc}
\cos n \theta_{j,i} & -\sin n \theta_{j,i} \\
\sin n \theta_{j,i} & \cos n \theta_{j,i}
\end{array}\right], O = \left[\begin{array}{cc}
0 & 0 \\
0 & 0
\end{array}\right].$}
\end{equation}
Here, $\theta_{j,i} = B^{-2i/d}_{j}, i \in [0,\cdots,d/2-1]$, is termed as the rotary angle of RoPE, and $B_j$ is typically a fixed base. 
The subscript $j$ serves to differentiate various RoPE angles $\Theta$, each associated with a distinct $B_j$, a distinction necessary for discussions in Section~\ref{sec:method}.
This approach effectively incorporates relative position information between $m$ and $n$ in the query-key product during attention computation:
\begin{equation}
\mathbf{\hat{q}}_{n}^{h\top}\cdot\mathbf{\hat{k}}^{h}_m=\left(\mathbf{R}_{\Theta_j, n} \cdot\mathbf{q}^{h}_n\right)^{\top}\left(\mathbf{R}_{\Theta_j, m}\cdot \mathbf{k}^{h}_m\right)=\mathbf{q}_n^{h\top} \cdot\mathbf{R}_{\Theta_j, m-n} \cdot\mathbf{k}^{h}_m,
\end{equation}
\begin{equation}
\label{eq:attn_rope}
\mathbf{Attn}^{h}_{nm}=\texttt{Softmax}\left(
\frac{\mathbf{\hat{q}}_{n}^{h\top}\cdot\mathbf{\hat{k}}^{h}_m}{\sqrt{d}}\right) = \texttt{Softmax}\left(
\frac{\mathbf{q}_n^{h\top} \cdot\mathbf{R}_{\Theta_j, m-n} \cdot\mathbf{k}^{h}_m}{\sqrt{d}}\right).
\end{equation}
Here, $\mathbf{Attn}^{h}_{nm}$ denotes the attention score assigned by the $h$-th head at position $n$ to position $m$.

\paragraph{Context awareness of LLMs}

\begin{wrapfigure}{r}{0.35\linewidth}
\vspace{-8mm}
\begin{center}
    \includegraphics[width=\linewidth]{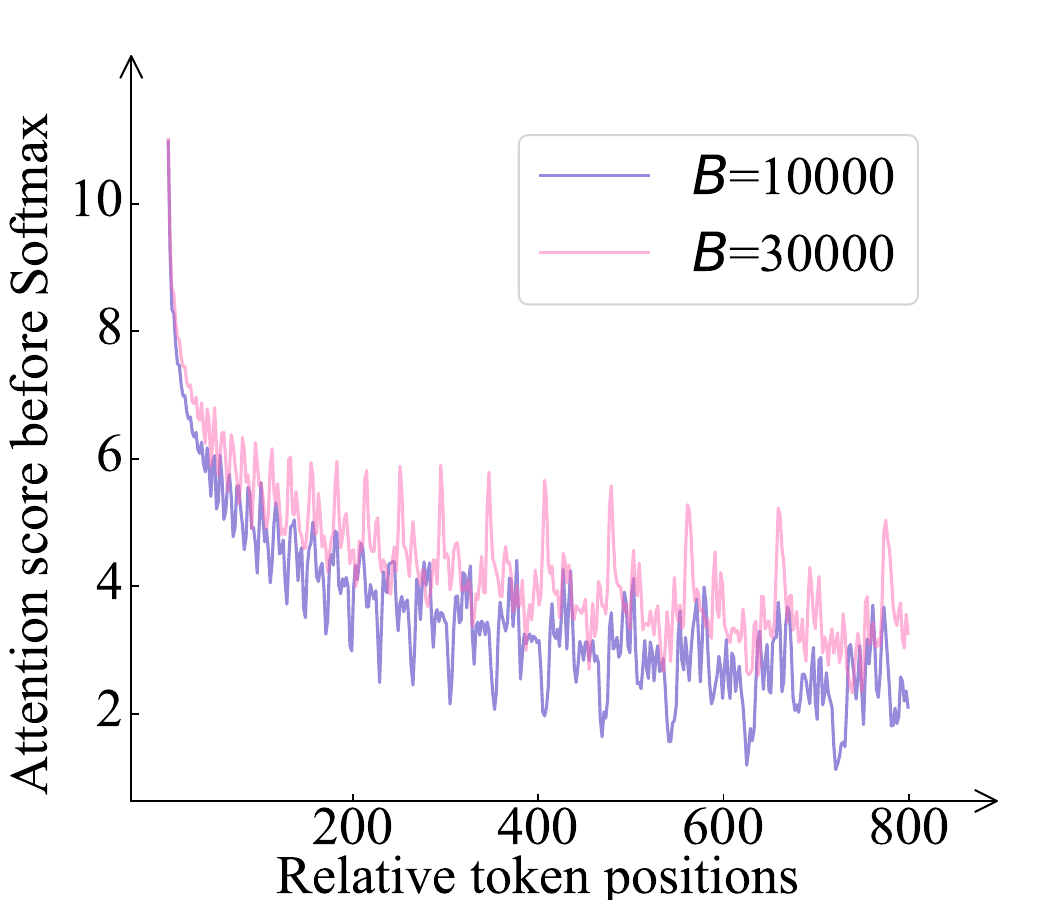}
    \end{center}
    \caption{Different $\Theta_j$ alter the upper bounds of attention scores between a token and its $x$-distance neighbors.
    Each angle is distinguished by its own base value $B_j$.}
    \label{fig:waveform}
    \vspace{-8mm}
\end{wrapfigure}
LLMs struggle with limited context awareness, significantly impacting their performance in tasks like long-text generation~\cite{zhang2024found}, Retrieval-Augmented Generation (RAG,~\cite{huang2024survey,chen2024benchmarking,cheng2024lift,tantao2024rag}), and multi-turn human-agent interactions~\cite{chen2023fortify} involving complex contexts. 
Liu et al.~\cite{liu2023lost} identified a problem known as ``Lost-in-the-Middle,'' where LLMs process the beginning and end of the context well but have reduced awareness of the middle. 
Chen et al.~\cite{chen2023fortify} observed that LLMs using RoPE exhibit uneven context awareness, favoring certain positions.
Peysakhovich et al.~\cite{peysakhovich2023attention} further highlighted that LLMs exhibit variable attention to document-level token segments based on their contextual positions.
Lv et al.~\cite{lv2024languagemodelsgrokcopy} observed that language models develop context awareness, especially in their ability to copy, through ``grokking~\cite{power2022grokkinggeneralizationoverfittingsmall}.''
They suggest pre-training models with increased regularization to enhance this capability.

\paragraph{Attention waveforms}
According to Chen et al.~\cite{chen2023fortify}, LLM's uneven awareness of different contextual positions is due to RoPE's mathematical characteristics.
Within RoPE, the attention score exhibits ``waveforms'' when retrieving the same token from the context, based on their relative positions. 
The troughs in these waveforms can impair task performance, especially when critical tokens are situated at these positions during generation.
Different RoPE angles produce waveforms with troughs occurring at different positions.
These phenomena are depicted in Figure~\ref{fig:waveform}.
A detailed derivation of the depicted curves in Figure~\ref{fig:waveform} is provided in Appendix~\ref{apx:attn-wave}.

\section{Mixture of In-Context Experts}
\label{sec:method}
In this section, we first introduce the core component of MoICE, the MoICE router, detailed in Section~\ref{sec:moice}. 
Subsequently, we delve into the optimization of MoICE in Section~\ref{sec:router-only-train}. 
Figure~\ref{fig:overview} provides an overview of MoICE.
The discussion in this section focuses solely on a single layer of transformer for clarity, with the same principles applying to any other layers.

\subsection{Architecture}
\label{sec:moice}
\begin{wrapfigure}{r}{0.4\linewidth}
\vspace{-5mm}
\begin{center}
    \includegraphics[width=\linewidth]{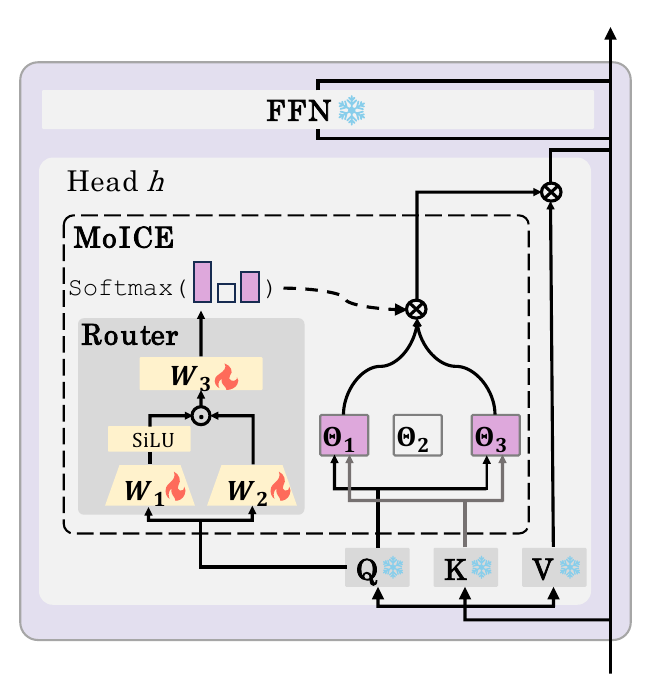}
    \caption{The structure of MoICE. 
    Only the router's parameters are trainable when plugged into an LLM. 
    For clarity, the figure illustrates a single head, with $N$=3 and $K$=2 as toy demonstration examples.}
    \label{fig:overview}
 \end{center}
\end{wrapfigure}
We aim to design an enhanced attention mechanism in LLMs that dynamically attends to crucial information across various contextual positions required for completing the head's function.
As a result, we can mitigate the performance drop caused by inadequate context awareness. 
Motivated by insights of Chen et al.~\cite{chen2023fortify}, who demonstrated that a distinct RoPE angle $\Theta_j$ could direct the attention heads more focus on specific contextual positions, we propose the integration of a contextual-aware routing mechanism.
This routing mechanism is designed to select the appropriate RoPE angles for processing a token. 
We implement the router as a Multi-Layer Perceptron (MLP) with the $\texttt{SiLU}$ activation function:
\begin{equation}
\label{eq:router}
\texttt{Router}\left(\mathbf{q}\right):=\mathbf{W}_{3} \left(\texttt{SiLU}\left(\mathbf{W}_{1}\mathbf{q}\right) \odot \left(\mathbf{W}_{2}\mathbf{q}\right)\right).
\end{equation}

Here, $\textbf{q}$ is the query vector that encapsulates the contextual information for the task.
This router input indicates the specific information for which the current token is ``querying.''
$\mathbf{W}_{1}, \mathbf{W}_{2} \in \mathbb{R}^{N \times d}$, and $\mathbf{W}_{3} \in \mathbb{R}^{N \times N}$ are weight matrices, where $N$ denotes the number of the number of RoPE angle candidates. 
Considering each head's distinct function~\cite{olsson2022context,wang2023interpretability,lv2024interpreting}, we integrate a router into every attention head in the LLM.
Notably, a router's decision is independent of other heads and dynamic to the context.

As defined in Eq.~\ref{eq:router}, the router outputs an $N$-tuple distribution, indicating the weight it allocates for each RoPE angle.
In each step in generation, the router selects $K$ angles from a set of $N$ angles $\{\Theta_1, \Theta_2,\ ......\ , \Theta_N\}$ for attention computation.
Specifically, we first identify the $K$ RoPE angles with the highest routing weights and normalize their relative weights using the \texttt{Softmax} function, resulting in $\mathbf{p}_n^{h} \in \mathbb{R}^{K}$, representing the probability distribution over the selected RoPE angles within the $h$-th head:
\begin{equation}
\begin{aligned}
&\text{TopK-Indices}_n^{h} = \texttt{argsort}(\texttt{Router}\left(\mathbf{q}_n^{h}\right))[:K],\\
&\textbf{p}_{n}^{h} = \texttt{Softmax}\left(\texttt{Router}\left(\mathbf{q}_n^{h}\right)[\text{TopK-Indices}_n^{h}]\right),
\end{aligned}
\label{eq:topk}
\end{equation}
where $\mathbf{q}_n^{h}$ represents the query at position $n$ within the $h$-th attention head in a Transformer layer.
Subsequently, we aggregate the attention scores computed with these chosen $K$ in-context experts based on their routing weights to derive the final attention scores for head $h$ from position $n$ to position $m$:
\begin{equation}
\mathbf{Attn}_{nm}^{h}=\sum_{j \in \text{TopK-Indices}^{h}_{n}} \mathbf{p}_n^{h}[j]\cdot
\texttt{Softmax}\left(
\frac{\mathbf{q}_n^{h\top} \cdot\mathbf{R}_{\Theta_j, m-n} \cdot\mathbf{k}^{h}_m}{\sqrt{d}}\right).
\end{equation}

Considering RoPE angles impact how attention heads allocate attention and focus on specific contextual positions, we view each distinct RoPE angle as an in-context expert, in contrast to traditional in-weight experts~\cite{fedus2022switch,jiang2024mixtral,dai2024deepseekmoe,gong2024mixtureofmodulesreinventingtransformersdynamic}, where the experts are learnable parameter weights.
Given that these in-context experts together augment LLMs' context awareness, we term this method Mixture of In-Context Experts (MoICE, Section~\ref{sec:method}).
Figure~\ref{fig:overview} illustrates the overview of MoICE. 
Our proposed MoICE has three major advantages:

% 不要提ab了，直接说只在qk乘法做了额外开销，实际增长的现存非常少，并且对推理速度影响非常小（Section xx）。真别搞chatgpt了，superfluous我这辈子第一次见
(1) We only add additional computational overhead to the query-key dot products, resulting in a minimal increase in memory usage and a negligible impact on inference speed (Section~\ref{sec:leval}).

(2) MoICE dynamically selects suitable RoPE angles token-wise and head-wise, offering unprecedented flexibility and unlocking the full potential of each attention head.

(3) Concerning LLMs' context awareness enhancement, MoICE addresses a longstanding issue: the relative position of the relevant information will shift during generation, leading to previous static modification of the attention heads~\cite{chen2023fortify,zhang2024enhancing} will be sub-optimal during practical generation. The contextual-aware dynamic routing in MoICE is not bothered by this issue.

\subsection{Router-only training}
\label{sec:router-only-train}

%To train the newly incorporated MoICE routers in LLMs, two choices are feasible: (1) simultaneously updating the LLMs' parameters alongside the routers, or (2) freezing the LLMs' parameters and solely optimizing the routers.

%Our experiments show that fine-tuning all LLM parameters alongside the routers leads to catastrophic forgetting. In contrast, training only the router with minimal steps is effective and efficient, as detailed in Section \ref{sec:exp}.
%Hence, we propose the router-only training strategy.

%\textcolor{red}{To train the newly incorporated MoICE routers in LLMs, the most straightforward way is to simultaneously update the LLMs' parameters alongside the routers. However, updating the original LLMs' parameters can result in catastrophic forgetting. To solve this, we propose the router-only training strategy, which freezes the LLMs' parameters and solely optimizing the routers. As detailed in Section \ref{sec:exp}, training only the router with minimal steps is effective and efficient.}
To train the newly incorporated MoICE routers in LLMs, the most straightforward way is to simultaneously update the LLMs' parameters alongside the routers. However, updating the original LLMs' parameters can result in catastrophic forgetting.  Therefore, we propose a more effective and efficient strategy, the router-only training strategy, which freezes the LLMs' parameters and solely optimizing the routers.

Given an input sequence, we calculate the negative log-likelihood loss ($\mathcal{L}_{nll}$) for language modeling. 
During backward propagation, only the parameters of MoICE routers are updated.
To mitigate the possibility of a router favoring specific experts disproportionately~\cite{fedus2022switch,wang2024graph}, we incorporate an auxiliary loss $\mathcal{L}_{aux}$ following~\cite{fedus2022switch}.
An ablation study on this auxiliary loss is in Table~\ref{tab:aux_loss}.
Given an input of $T$ tokens and $N$ experts, we calculate the $\mathcal{L}_{lb}$ by the scaled dot-product between frequency vector $\mathbf{F}$ and probability vector $\mathbf{P}$:
\begin{equation}
    \mathcal{L}_{aux}= \alpha \cdot N \cdot \sum_{j=1}^{N} \mathbf{F}_j \cdot \mathbf{P}_j.
    \label{eq:balance}
\end{equation}
Eq.~\ref{eq:balance} avoids the router falling into a sub-optimal solution favoring specific experts overwhelmingly, as its minimal is achieved when the routing probability is uniform.
Here, $\alpha$ is the weighting factor for load balancing loss. $\mathbf{F}_j$ denotes the proportion of the $j$-th expert selected across all positions and attention heads, while $\mathbf{P}_j$ denotes the proportion of router weight assigned to expert $j$:
\begin{equation}
\begin{aligned}
    &\mathbf{F}_j=\frac{1}{T \times H} \sum_{t=1}^{T} \sum_{h=1}^{H} \mathbbm{1}\{j \in  
 \text{TopK-Indices}_t^{h} \},\\
    &\mathbf{P}_j=\frac{1}{T \times H} \sum_{t=1}^{T} \sum_{h=1}^{H} \mathbbm{1}\{j \in  
 \text{TopK-Indices}_t^{h} \} \cdot \textbf{p}^{h}_{t}[j].
\end{aligned}
\end{equation}
Our overall training objective is to minimize the following loss: 
\begin{equation}
    \label{eq:all_loss}
    \mathcal{L} = \mathcal{L}_{nll} + \mathcal{L}_{aux}.
\end{equation}

\section{Experiment}
\label{sec:exp}
\subsection{Setup}
To evaluate the efficacy of MoICE, we implement it with open-source LLMs, which we will introduce later, and conduct lightweight training of MoICE routers on a small and general dataset.
Subsequently, we evaluate the enhanced LLM's capability to zero-shot undertake multiple tasks in long context understanding and generation, as detailed in Section~\ref{sec:leval} and Section~\ref{sec:rag}.

\paragraph{Training data} 
We use a training dataset\footnote{\url{https://huggingface.co/datasets/HuggingFaceH4/OpenHermes-2.5-1k-longest}} which extracts the one thousand longest entries from OpenHermes~\cite{OpenHermes}.
OpenHermes is a multi-source integrated dataset containing high-quality synthetically generated instruction and chat samples. 
A detailed analysis of other training data is in Section~\ref{sec:data}.

\paragraph{Hyperparameters for MoICE-router-only training} 

We froze all the original parameters of the open-source LLMs we used and only trained the MoICE router.
Following Attention Buckets~\cite{chen2023fortify}, we employed the RoPE angle set of $N=7$ items, each assigned with base values as follows: $\{10,000, 17,500, 18,000, 19,000, 20,000, 22,500, 25,000\}$.
By default, unless otherwise specified, the attention head selects $K$=7 bases to ensure a fair comparison with~\cite{chen2023fortify}. 
Section~\ref{sec:leval} introduces our baselines in detail.
Section~\ref{sec:base-search} delves into the impact of set size and the number of selected items.

We implement a warm-up strategy comprising 20\% of the total steps, with a maximum learning rate of 0.0001.
The batch size is 128.
$\alpha$ is set as 0.3.
We train the MoICE routers for 1 epoch (about 8 minutes) on four A800-80G GPUs.

\subsection{Long context understanding and generation}
\label{sec:leval}
Following the L-Eval benchmark~\cite{an2023eval}, we evaluated the LLM with tasks categorized into two main groups: closed-ended and open-ended tasks.
Closed-ended tasks primarily focus on the capacity for understanding and reasoning within long contexts,  including tasks like multiple-choice questions from QuALITY~\cite{bowman2022quality}, Coursera,~\footnote{\url{https://coursera.org/}} TOEFL~\cite{chung2017supervised}, and True/False question answering from SFiction.~\footnote{\url{https://github.com/nschaetti/SFGram-dataset}}
On the other hand, open-ended tasks include summarization generation and open-format question-answering tasks, requiring extracting information from lengthy in-context documents. 
The open-ended tasks comprise a subset of 181 questions drawn from 29 diverse long documents.

\paragraph{Baselines and open-source LLMs} 
In evaluating the efficacy of our proposed MoICE, we compare it against several state-of-the-art methods known for enhancing the capacity of LLMs to understand and generate long contexts.
These baselines include two context extrapolation techniques: Positional Interpolation (PI,~\cite{chen2023extending}) and Dynamic NTK~\cite{ntk}. 
Additionally, we consider two inference algorithms for context-awareness enhancement: Ms-PoE~\cite{zhang2024found} and Attention Buckets~\cite{chen2023fortify}. 

We evaluate all these methods alongside our MoICE on three representative open-source LLMs that utilize RoPE for positional embeddings: Llama2-7B-Chat~\cite{touvron2023llama}, Mistral-7B-Instruct-v0.1~\cite{jiang2023mistral} and Qwen1.5-7B-Chat~\cite{qwen}.
Llama2-7B and Qwen1.5-7B support a pre-trained context length of 4,096 and 32,768, respectively.
Mistral-7B employs a sliding window attention (SWA) mechanism with a window size of 4,096 tokens, enabling it to accommodate longer contexts than the default. 
Therefore, we conduct experiments with a context length of 8,192 on Mistral-7B, using SWA as the exclusive baseline for comparison.
For PI and Dynamic NTK, we apply a scaling ratio of 1.5, while for the remaining baselines, we adhere to the hyperparameters specified in their original papers.
All methods are tested on a single A800-80G GPU, except for applying AB to Mistral-7B-8k, which needs 2 GPUs due to substantial memory requirements.

% \usepackage{booktabs}
% \usepackage[normalem]{ulem}
\begin{table}[t]
\caption{Experimental results on the L-Eval Benchmark~\cite{an2023eval}. 
Applying to various models, MoICE demonstrate superior performance compared to previous competitive approaches.
We emphasize the highest score in bold.}
\label{table:Leval}
\resizebox{\linewidth}{!}{
\begin{threeparttable}
\begin{tabular}{@{}lcccccccc@{}}
\toprule
\multirow{2}*{\textbf{Method}} & \multicolumn{5}{c}{\textbf{Closed - Ended Task}} & \multicolumn{3}{c}{\textbf{Open - Ended Task}}\\ \cmidrule(r){2-6} \cmidrule(r){7-9}
                     ~ & \multicolumn{1}{c}{\textbf{Coursera}} & \multicolumn{1}{c}{\textbf{QuALITY}} & \multicolumn{1}{c}{\textbf{TOEFL}} & \multicolumn{1}{c}{\textbf{SFiction}} & \multicolumn{1}{c}{\textbf{Average}} & \multicolumn{1}{c}{\textbf{wins}} & \multicolumn{1}{c}{\textbf{ties}} & \multicolumn{1}{c}{\textbf{win-rate\%}\tnote{*}} \\ \midrule
Llama2-7B-Chat~\cite{touvron2023llama}              & 36.77                        & 38.12                       & 55.02                     & 60.16                        & 47.52                       & 68                       & 117                      & 34.94                        \\\midrule
\ + Fine-tuning & 32.85                        & 30.20                       & 51.30                     & 59.38                        & 43.43                       & 65                       & 91                       & 30.52                        \\
\ + PI~\cite{chen2023extending}             & 38.23                        & 38.61                       & 56.51              & 61.72                        & 48.77                       & 76                       & 112                      & 36.46                        \\
\ + Dynamic NTK~\cite{ntk}             & 40.26                        & 39.11                       & 55.76                     & 62.50                        & 49.41                       & 82                       & 112                      & 38.12                        \\
\ + Ms-PoE~\cite{zhang2024found}             & 39.24                        & 40.10                       & 55.76                     & 63.28               & 49.60                       & 86                 & 110                      & 38.95                        \\
\ + AB~\cite{chen2023fortify}         & \textbf{40.41}               & 41.09             & \textbf{56.88}                     & 61.72                        & 50.02                 & 85                       & 114               & 39.23                  \\
\ + MoICE (Ours)                  & 39.83               & \textbf{42.08}                 & 56.13            & \textbf{64.84}               & \textbf{50.72}              & \textbf{89}              & \textbf{118}             & \textbf{40.88}               \\ \midrule
\midrule 
Mistral-7B-Instruct-8k~\cite{jiang2023mistral}         & 45.20 & 44.06 & 62.08 & 61.72 & 53.27 & 71 & 105 & 34.11 \\\midrule
\ + Fine-tuning  & 25.29 & 26.73 & 25.65 & 50.00 & 31.92 & 53 & 85 & 26.38 \\
\ + SWA         & 44.77 & 42.57 & 62.08 & 60.94 & 52.59 & 73 & 89  & 32.45 \\
\ + PI~\cite{chen2023extending}  & 44.19 & 44.06 & \textbf{64.68} & 62.50 & 53.86 & 73 & 96  & 33.43 \\
\ + Dynamic NTK~\cite{ntk} & 45.35 & 42.08 & 62.08 & \textbf{63.28} & 53.20 & 78 & 103 & 35.77 \\
\ + Ms-PoE~\cite{zhang2024found} & 46.37 & 45.05 & 61.34 & 57.03 & 52.45 & 84 & 106 & 37.84 \\
\ + AB~\cite{chen2023fortify}  & 46.08 & 42.57 & 62.08 & 62.50 & 53.31 & \textbf{87} & 110 & 39.22 \\
\ + MoICE (Ours)  & \textbf{47.82} & \textbf{46.53} & \textbf{64.68} & 62.50 & \textbf{55.38} & 85 & \textbf{117} & \textbf{39.36} \\ \midrule \midrule
Qwen1.5-7B-Chat~\cite{qwen} & \textbf{78.44} & 61.88 & 61.19 & 69.53 & 67.76 & 83 & \textbf{119} & 40.83 \\ \midrule
\ + PI~\cite{chen2023extending} & 76.58 & 61.88 & 60.32 & 70.31 & 67.27 & 83 & 107 & 39.11 \\ 
\ + Dynamic NTK~\cite{ntk} & 78.07 & 62.38 & 60.32 & 70.31 & 67.77 & 84 & 111 & 40.20 \\ 
\ + Ms-PoE~\cite{zhang2024found} & 75.47 & 60.89 & 60.47 & \textbf{71.88} & 67.18 & OOM & OOM & N/A \\ 
\ + AB~\cite{chen2023fortify} & \textbf{78.44} & OOM & OOM & OOM & N/A & OOM & OOM & N/A \\
\ + MoICE (Ours) & \textbf{78.44} & \textbf{62.87} & \textbf{61.77} & 71.09 & \textbf{68.54} & \textbf{91} & 105 & \textbf{41.59} \\

\bottomrule
\end{tabular}
\begin{tablenotes}
      \item[*] Following~\cite{an2023eval}, win-rate = (win counts + 0.5 * tie counts)
\end{tablenotes}
\end{threeparttable}}
\end{table}

\begin{table}[t]
\centering
\caption{Practical inference time (in minutes) / GPU memory costs (GB) on a single A800-80G GPU for each method applied to Llama2-7B-Chat (top) and Mistral-7B-Instruct-8k (bottom), respectively.
Due to out-of-memory issues, AB can not accomplish many tasks, denoted as OOM in the table.}
\label{tab:time-cost}
\resizebox{0.85\linewidth}{!}{
\begin{tabular}{@{}lcccccc@{}}
\toprule
\textbf{Method} & \textbf{Coursera $\downarrow$} & \textbf{QuALITY $\downarrow$} & \textbf{TOEFL $\downarrow$} & \textbf{SFiction $\downarrow$} & \textbf{Open-Ended $\downarrow$} & \textbf{Average $\downarrow$} \\ \midrule
AB~\cite{chen2023fortify} & 10.9 / 78.7 & 18.1 / 62.5 & 19.9 / 56.5 & 5.0 / 33.2 & 45.9 / 78.2 &  20.0 / 61.8 \\
Ms-PoE~\cite{zhang2024found} &  4.1 / 27.2 & 6.0 / 27.8 & 6.7 / 28.6 &  6.0 / 27.8 & 20.2 / 28.9 & 8.6 / 28.1 \\
MoICE (Ours) & 5.0 / 19.6 & 11.0 / 19.7 &  10.2 / 19.5 & 1.6 / 15.2 & 34.2 / 23.2 & 12.4 / 19.4 \\
\midrule
\midrule
AB~\cite{chen2023fortify} & OOM & OOM & 37.2 / 71.4 & OOM & OOM & N/A \\
Ms-PoE~\cite{zhang2024found} & 14.1 / 50.3 & 11.2 / 48.4 & 9.8 / 25.4 & 4.5 / 50.3 & 72.8 / 62.4 & 22.5 / 47.4 \\
MoICE (Ours)  & 13.4 / 25.7 & 7.7 / 22.9 & 11.3 / 20.4 & 2.3 / 22.8 & 77.8 / 29.3 & 22.5 / 24.2 \\\bottomrule
\end{tabular}}
\end{table}

\paragraph{Evaluation metrics} We adopt the exact match for closed-ended tasks. 
For open-ended tasks, we employ \textit{GPT-4-Turbo}~\cite{openai2024gpt4} as the judge to evaluate the effectiveness of various enhancement methods on open-source LLMs. 
This evaluation compares their performance against \textit{GPT3.5-Turbo-16k-0613} across 181 questions.

\paragraph{Results and analysis}
We report our experimental results in Table~\ref{table:Leval}. 
MoICE significantly enhances the overall performance of Llama-2-7B-chat (with p-value < 0.02 in the t-test) in both closed-ended and open-ended tasks. On Mistral, MoICE outperforms all baseline models significantly (p-value < 0.02).
We also report the mean and standard deviation of MoICE in Table~\ref{table:mean}.
Standard fine-tuning degrades the performance of original LLMs, demonstrating catastrophic forgetting and proving that the improvement of MoICE does not stem from more training.
These results underscore MoICE's efficacy in enhancing LLMs' ability to understand and generate long contexts, both of which require high context awareness. Furthermore, these results underscore the broad applicability of MoICE across different LLMs.

Regarding efficiency, we provide practical inference time and memory costs associated with AB, Ms-PoE, and MoICE in Table~\ref{tab:time-cost}. 
For a fair comparison, we utilize Flash Attention 2~\cite{dao2023flashattention2} across all approaches. While achieving superior overall performance, MoICE remains at an inference speed similar to Ms-PoE and notably excels in memory efficiency compared to these two baselines.

We also perform further experiments on one additional long context benchmark LongBench~\cite{bai2023longbench}, which are detailed in Appendix~\ref{apx:longbench}.

\subsection{Retrieval-augmented generation (RAG)}
\label{sec:rag}
Retrieval-augmented generation (RAG) tasks involve retrieving numerous documents related to the current generation.
The retrieved documents are arranged in the context.
RAG necessitates that LLMs have robust context awareness to pinpoint crucial documents, process the retrieved information effectively, and integrate it to generate responses.

Following~\cite{chen2023fortify,zhang2024found}, we employ the MDQA task to evaluate the efficacy of MoICE in enhancing LLMs' performance in RAG tasks.
Meanwhile, MDQA offers the bonus of allowing flexible control over the location of documents, enabling a more precise evaluation of LLMs' context awareness across various contextual positions.

\begin{table}[t]
\caption{The experiment results on the MDQA task. MoICE achieve superior average performance compared to previous competitive approaches. We emphasize the highest score in bold.
}
%In the MDQA task, MoICE consistently outperforms baselines regardless of the position of the relevant document within the context. It notably enhances the performance of LLMs and achieves minimal gap between the highest and lowest accuracy scores across various positions.
\centering
\label{table:nih}
\resizebox{0.7\linewidth}{!}{
\begin{tabular}{@{}lccccccc@{}}
\toprule
\textbf{Method}         & \textbf{1}              & \textbf{3}              & \textbf{5}              & \textbf{7 }             & \textbf{10}   & \textbf{Gap}         & \textbf{Avg.}            \\
\midrule
Llama2-7B-Chat & 64.14          & 65.95   & 64.97        & 62.67    & \textbf{67.53} & 4.86 & 65.05          \\\midrule
% \ + PI~\cite{chen2023extending}             & 65.27          & 65.46          & 65.08    & 62.71          & 65.01         & {2.75} & 64.71          \\
\ + Ms-PoE~\cite{zhang2024found}          & {66.06}   & 64.29          & 63.99          & 62.22          & 64.75         &3.84  & 64.34          \\
\ + AB~\cite{chen2023extending} & \textbf{66.36} & 66.14 & {65.25} & {63.20} & 64.93 & 3.16 & {65.18} \\
\ + MoICE (Ours) & 65.50 & \textbf{66.33} & \textbf{65.61} & \textbf{64.11} & {65.84}  &\textbf{2.22}  & \textbf{65.48} 
\\ 
\midrule\midrule
\textbf{Method}         & \textbf{1}              & \textbf{8}              & \textbf{15}              & \textbf{23 }             & \textbf{30}   & \textbf{Gap}         & \textbf{Avg.}            \\
\midrule
% Mistral-7B-Instruct-8k & 57.33 & 52.20
%  & 57.66 & 58.92 & 60.18 &- &57.26 \\\midrule
% \ + PI  & 53.10 & 46.40 & 46.78 & 44.37 & 45.73&- & 47.28  \\
% \ + Ms-PoE~\cite{zhang2024found} & 51.07 & 48.44 & 48.66 & 50.06 & 
% 52.20 &- & 50.09 \\
% \ + AB~\cite{chen2023extending} & 57.89 & 54.05 & 56.08 & 58.00 & 59.84 &- & 57.17  \\
% \ + AB2(ours) & 59.96 & 54.46 & 56.69 & 54.58 & 55.44 &- & 56.22
%  \\\bottomrule
Mistral-7B-Instruct-8k & 58.38 & 47.42 & 46.97 & 49.68 & 50.81 &\textbf{11.41}&50.65 \\\midrule
% \ + PI  & 51.93 &	42.90	& 39.32	& 36.95	&38.19 &14.98&41.86 \\
\ + Ms-PoE~\cite{zhang2024found} &  52.76&41.24&42.80&42.90&43.58&11.52&44.66\\
\ + AB~\cite{chen2023extending} & 58.57 &	47.57	&47.12	&49.83	&\textbf{50.96}& 11.45&50.81 \\
\ + MoICE (Ours) & \textbf{61.81} &\textbf{52.54}&\textbf{52.43}&\textbf{50.36}&49.34 &12.47&\textbf{53.30}
 \\\bottomrule
\end{tabular}}
\end{table}

Our MDQA experiments leverage a subset of NaturalQuestions-Open~\cite{lee-etal-2019-latent,kwiatkowski-etal-2019-natural}, consisting of 2,655 queries, following~\cite{zhang2024found,liu2023lost}.
Each query is paired with a context consisting of 10 or 30 documents (with an average of 1,722 or 5,046 tokens), depending on the model (Llama-2-7B-chat or Mistral-7B-Instruct-8k), tasked with answering based on this contextual information. 
Only one document among these comprises useful information for the given query. 
We compare Ms-PoE, AB, and MoICE, testing each method through 5 iterations. 
For Llama, the relevant document is positioned 1st, 3rd, 5th, 7th, and 10th within the context, while for Mistral, it is positioned 1st, 8th, 15th, 23rd, and 30th, respectively.

In Table~\ref{table:nih}, MoICE on Llama demonstrates the highest average performance across most positions, showcasing its remarkable stability. 
Its accuracy scores show minimal variation, with only a marginal difference of 2.22 points between its highest and lowest values. 
On Mistral, MoICE exhibits significant average improvement (p-value < 0.02). 
Notably, when the relevant document is positioned at the end of the context, all methods on Llama exhibit a decrease compared to the original model, although MoICE shows a minimal decline.
This phenomenon also happens in the Mistral model. 
We posit that this decline may stem from the original model predominantly directing attention towards nearest documents~\cite{liu2023lost,peysakhovich2023attention}. However, as approaches enhance awareness of various contextual positions, the model's attention to the nearest documents is diffused by other positions, as its overall capacity for context awareness is constant and limited. 
Nevertheless, MoICE consistently emerges as the superior-performing method overall across language models.

\section{Method analysis}
\label{sec:base-search}
In this section, we delve into a comprehensive analysis of the properties of MoICE. 
We illustrate how $N$, the total number of in-context experts (Section~\ref{sec:n}), and $K$, the specific number of selected in-context experts (Section~\ref{sec:k}), influence MoICE. We further demonstrate that MoICE is robust to training data  (Section~\ref{sec:data})
Additionally, we present a case study demonstrating the dynamic selection of in-context experts for tokens during generation (Section~\ref{sec:dynamic}). Finally, we verify the effectiveness of language model with MoICE architecture in pretraining stage (Section~\ref{sec:pretraining}).

\subsection{The effect of expert total numbers \texorpdfstring{$N$}{}}
\label{sec:n}
We investigate the impact of the total number of experts. 
Employing the search algorithm proposed by Chen et al.~\cite{chen2023fortify}, we obtain various sets of different sizes, each comprising complementary base values. 
The searched expert sets are detailed in Appendix~\ref{apx:base-set}.
We apply MoICE to Llama-2-7B-chat and test the model on L-Eval tasks.
The results are presented in Table~\ref{tab:n}.
%Because the original Llama utilizes only one expert, its results are denoted as ($N$=1) in the table. 
The results of the original Llama are denoted as ($N$=1) in the table.
As the table illustrates, MoICE demonstrates improvement to LLMs' context awareness with increasing $N$, with noticeable improvements even when $N$ is as small as 3. 
However, as $N$ reaches 9, the average performance is close to $N$=7, indicating a performance plateau. 
This suggests that having $N$=7 experts is sufficient for general usage.

\subsection{The effect of selected experts number \texorpdfstring{$K$}{}}
\label{sec:k}
With a fixed number of experts ($N$=7), we examine the effect of different numbers of chosen experts $K$ with values of 1, 3, 5, and 7. 
We consider two additional setups where 7 experts are selected with equal weights (``Equal Weights'') and with random weights (``Random Weights''  ), using Llama-2-7B-chat as the case study.
The results are presented in Table~\ref{tab:k}.
Setting $K=1$ doesn't enhance or significantly degrade the model's performance, aligning with our assertion in the Introduction (Section~\ref{sec:intro}): assigning a single and unique RoPE angle to each head inadequately explores the head's functionality.
For $K$ greater than 3, performance improvements become evident. This shows that the MoICE router in our method can select the appropriate combination of experts to better aware the context.
Randomly selecting experts ruins the model's language modeling ability, leading to aberrant outputs.
%

% \usepackage{booktabs}
\begin{table}[t]
\centering
\caption{The performance of Llama-2-7B-chat enhanced by MoICE with $N$ in-context experts. We show results marked with color to emphasize the improvements over the original model. }
\label{tab:n}
\resizebox{0.65\linewidth}{!}{
\begin{threeparttable}
\begin{tabular}{@{}cccccc}
\toprule
    \textbf{Method} & \textbf{Coursera} & \textbf{QuALITY} & \textbf{TOEFL} & \textbf{SFiction} & \textbf{Avg.}   \\\midrule
Original ($N$=1) & 36.77 & 38.12 & 55.02 & 60.16 & 47.52 \\\midrule
$N$=3          &\cellcolor{purple!8}{37.65} &\cellcolor{purple!8}{40.10} & \cellcolor{purple!8}{55.76} & \cellcolor{purple!8}{62.50}& \cellcolor{purple!8}{49.00} \\
$N$=5          &\cellcolor{purple!8}{38.23} & \cellcolor{purple!8}{39.60} & \textbf{\cellcolor{purple!8}{56.13}} & \cellcolor{purple!8}{63.28} & \cellcolor{purple!8}{49.32} \\
$N$=7          & \cellcolor{purple!8}{39.83}  & \textbf{\cellcolor{purple!8}{42.08}} & \textbf{\cellcolor{purple!8}{56.13}} & \textbf{\cellcolor{purple!8}{64.84}} & \textbf{\cellcolor{purple!8}{50.72}} \\
$N$=9          & \textbf{\cellcolor{purple!8}{40.26}} & \cellcolor{purple!8}{41.58} & \textbf{\cellcolor{purple!8}{56.13}} & \textbf{\cellcolor{purple!8}{64.84}} & \cellcolor{purple!8}{50.70} \\\bottomrule
\end{tabular}
\end{threeparttable}}
\end{table}

\begin{table}[t]
\centering
\caption{The improvement of context awareness of Llama-2-7B-chat by MoICE, wherein each head dynamically selects diverse $K$ experts ($N$=7). We show results marked with color to emphasize the improvements over the original model.}
\label{tab:k}
\resizebox{0.65\linewidth}{!}{
\begin{threeparttable}
\begin{tabular}{@{}cccccc}
\toprule
    \textbf{Method} & \textbf{Coursera} & \textbf{QuALITY} & \textbf{TOEFL} & \textbf{SFiction} & \textbf{Avg.}   \\\midrule
Original ($N$=1) & 36.77 & 38.12 & 55.02 & 60.16 & 47.52 \\\midrule
$K$=1          & 35.03 & 35.64 & \cellcolor{purple!8}{\textbf{56.51}} & \cellcolor{purple!8}{61.72} & 47.22 \\
$K$=3 &\cellcolor{purple!8}{\textbf{39.83}} & \cellcolor{purple!8}{41.58} &\cellcolor{purple!8}{56.13} & \cellcolor{purple!8}{\textbf{64.84}}& \cellcolor{purple!8}{50.60}\\
$K$=5 &\cellcolor{purple!8}{38.52} & \cellcolor{purple!8}{39.60} & \cellcolor{purple!8}{56.13} & \cellcolor{purple!8}{\textbf{64.84}} & \cellcolor{purple!8}{49.77}\\
$K$=7 & \cellcolor{purple!8}{\textbf{39.83}}  & \cellcolor{purple!8}{\textbf{42.08}} & \cellcolor{purple!8}{56.13} & \cellcolor{purple!8}{\textbf{64.84}} & \cellcolor{purple!8}{\textbf{50.72}} \\
% Equal Interval & \cellcolor{purple!8}{39.39} &\cellcolor{purple!8}{40.59} & \cellcolor{purple!8}{57.62} & \cellcolor{purple!8}{62.50} &  \cellcolor{purple!8}{50.03} \\
% Airoboros & \cellcolor{purple!8}{39.68} & \cellcolor{purple!8}{41.58} & \cellcolor{purple!8}{56.13} & \cellcolor{purple!8}{64.84} & \cellcolor{purple!8}{50.56} \\
Equal Weights & 36.48    & 38.12   & 53.90 & \cellcolor{purple!8}{61.72}    & \cellcolor{purple!8}{47.56} \\
Random Weights & 15.55 & 28.71  & 21.75  & 8.59  & 18.65    \\ 
\bottomrule
\end{tabular}
\end{threeparttable}
}
\end{table}

% \begin{figure}[t]
%     \centering
%     \includegraphics[width=\linewidth]{Styles/N_and_K.pdf}
%     \caption{The visualization of (a) the effect of expert total numbers $N$ and (b) selected experts number $K$. ``Original'' refers to the result of Llama-2-7B-chat~\cite{touvron2023llama}}
%     \label{fig:N_and_K}
%     \vspace{-3mm}
% \end{figure}

\subsection{MoICE is robust to training data}
\label{sec:data}
We further analyze the impact of the data for training routers. 
We additionally use three instruction fine-tuning datasets from different sources: a self-instruct dataset, Airoboros~\cite{airoboros}; and two datasets for LLM alignment with long context, Long-Alpaca~\cite{long-alpaca}, and LongAlign~\cite{bai2024longalign}.
The hyperparameters remain consistent as mentioned in Section~\ref{sec:exp}.
As presented in Table~\ref{tab:data1}, MoICE exhibits almost identical scores when trained on different data, showcasing the robustness of our method.

%\begin{table}[t]
%\label{tab:more_ablation}
%\centering
%\caption{The improvement of context awareness of Llama-2-7B-chat by %MoICE, we conduct more analysis experiments on RoPE base set and %gate-only training dataset.}
%\resizebox{0.75\linewidth}{!}{
%\begin{threeparttable}
%\begin{tabular}{@{}cccccc@{}}
%\toprule
%    \textbf{Method} & \textbf{Coursera} & \textbf{QuALITY} & %\textbf{TOEFL} & \textbf{SFiction} & \textbf{Avg.}   \\\midrule
%Original ($N$=1) & 36.77 & 38.12 & 55.02 & 60.16 & 47.52 \\\midrule
%Ours & \cellcolor{purple!8}{\textbf{39.83}}  & \cellcolor{purple!8}%{\textbf{42.08}} & \cellcolor{purple!8}{56.13} & %\cellcolor{purple!8}{\textbf{64.84}} & \cellcolor{purple!8}%{\textbf{50.72}} \\
%Base\_1 & 39.39 &40.59 & 57.62 & 62.50 &  50.03 \\
%airboros & 39.68 & 41.58 & 56.13 & 64.84 & 50.56 \\

%%Base_1 [10000,12500,15000,17500,20000,22500,25000]
%%airboros 新微调数据集
%\bottomrule
%\end{tabular}
%\end{threeparttable}
%}
%\end{table}

\begin{table}[!h]
\centering
\caption{The improvement of context awareness of Llama-2-7B-chat by MoICE trained on various data.}
\label{tab:data1}
\resizebox{0.7\linewidth}{!}{
\begin{threeparttable}
\begin{tabular}{@{}cccccc@{}}
\toprule
    \textbf{Training Data} & \textbf{Coursera} & \textbf{QuALITY} & \textbf{TOEFL} & \textbf{SFiction} & \textbf{Avg.}   \\\midrule
OpenHermes~\cite{OpenHermes} & 39.83 & 42.08 & 56.13 & 64.84 & 50.72 \\
% Equal Interval & 39.39} &40.59} & 57.62} & 62.50} &  50.03} \\
Airoboros~\cite{airoboros} & 39.68 & 41.58 & 56.13 & 64.84 & 50.56 \\
Long-Alpaca~\cite{long-alpaca} & 39.68 & 42.08 &56.13&64.84 & 50.68 \\
LongAlign~\cite{bai2024longalign} & 39.68 & 41.58 & 56.13 &64.84 &50.56 \\
\bottomrule
\end{tabular}
\end{threeparttable}
}
\end{table}
\subsection{The visualization of dynamic routing states}
\label{sec:dynamic}
We provide a case study exemplifying the dynamic routing mechanism within MoICE during text generation. 
Depicted in Figure~\ref{fig:router} in the Appendix, the MoICE router of each head independently selects distinct experts. 
At each step of the generation process, these heads dynamically choose experts for each new token. 
This dynamic utilization of diverse RoPE angles within each attention head maximizes the potential of attention heads across various inputs, a capability not attained in prior research, including both Attention Buckets and Ms-PoE.

\subsection{Applying MoICE to the pretraining stage}
\label{sec:pretraining}
We further evaluate the performance of a language model with MoICE architecture in pretraining stage.
Specifically, we train a language model with a Llama architecture of 49M parameters, with and without MoICE respectively. We pretrain a small model from scratch and observe the effectiveness of MoICE. 
More experimental details can be found in Appendix~\ref{apx:pretraining}.
We measure the model’s context awareness on the Key-Value Retrieval~\cite{liu2023lost} task, which uses multiple randomly generated key-value string pairs as prompts to evaluate the model's ability to extract the correct value corresponding to a given query from the context. One prompt example can be found in Figure~\ref{fig:kv_prompt}.

\begin{table}[h]
\centering
\caption{Experimental results on the Key-Value Retrieval task~\cite{bai2023longbench}, 
We evaluate the model's awareness of different positions by controlling the position index of the key-value pair corresponding to the query among 10 key-value pairs in the prompt.}
\label{tab:pretraining}
\resizebox{0.5\linewidth}{!}{
\begin{threeparttable}
\begin{tabular}{cccccc}
\toprule
    \textbf{Position} & \textbf{1} & \textbf{3} & \textbf{5} & \textbf{7} & \textbf{9} \\ \midrule
Baseline & 0.476 & 0.324 & 0.328 & 0.344 & 0.502 \\
+ MoICE & 0.652 & 0.762 & 0.634 & 0.622 & 0.814 \\ 
    
 \bottomrule   
    
\end{tabular}
\end{threeparttable}}
\end{table}

From the results in Table~\ref{tab:pretraining}, we can see that our model can significantly increase the contextual capabilities of the pretrained language model, which indicates the potential of scaling up our method in pretraining stage.

\section{Conclusion}

In this paper, we introduce a novel approach to enhancing the context awareness of LLMs termed \textit{Mixture of In-Context Experts} (MoICE). 
Through lightweight training, open-source LLMs such as Llama and Mistral, enhanced by MoICE, demonstrate improved context awareness. 
Across numerous tasks demanding substantial context awareness, MoICE-enhanced LLMs consistently outperform competitive baselines, all the while maintaining commendable efficiency.  
A distinctive feature of MoICE is that it first implements head- and token-specific RoPE angles assignment for attention heads, a pivotal factor contributing to its success.
This paper underscores the need to address the inherent limitations in current LLMs and advocates for a thorough exploration of their existing capabilities.

\begin{ack} 
This work was supported by the National Natural Science Foundation of China (NSFC Grant No. 62122089), Beijing Outstanding Young Scientist Program NO. BJJWZYJH012019100020098, and Intelligent Social Governance Platform, Major Innovation \& Planning Interdisciplinary Platform for the ``Double-First Class'' Initiative, Renmin University of China, the Fundamental Research Funds for the Central Universities, and the Research Funds of Renmin University of China.
Ang Lv is supported by the Outstanding Innovative Talents Cultivation Funded Programs 2024 of Renmin University of China.
\end{ack}

\bibliography{neurips_2024}
\bibliographystyle{plain}

\appendix

\section{Results on LongBench}
\label{apx:longbench}
LongBench~\cite{bai2023longbench} is a benchmark for bilingual, multitask, and comprehensive assessment of long context understanding capabilities of large language models. We choose 16 tasks from LongBench, spanning five long-text application scenarios: \textbf{Single-Doc QA} (NarrativeQA, Qasper, and MultiFieldQA-en), \textbf{Multi-Doc QA} (HotpotQA, 2WikiMQA and Musique), \textbf{Summarization} (GovReport, QMSum and MultiNews), \textbf{Few-shot Learning} (TREC, TriviaQA, SAMSum and LSHT) and \textbf{Synthetic Tasks} (Passage Count,	PassageRetrieval-en and PassageRetrieval-zh). 

For evaluation, we follow the setup of the LongBench benchmark. The input length of LLama2-7B is set to 4k, Mistral-7B to 8k, and Qwen1.5-7B to 32k.
We report the average task scores for each scenario for all methods in Table~\ref{tab:longbench}.

\begin{table}[h]
\centering
\caption{Experimental results on the LongBench Benchmark~\cite{bai2023longbench}. We emphasize the highest score in bold.}
\label{tab:longbench}
\resizebox{\linewidth}{!}{
\begin{threeparttable}
\begin{tabular}{@{}lcccccc@{}}
\toprule
    \textbf{Method} & \textbf{Single-Doc QA} & \textbf{Multi-Doc QA} &\textbf{ Summarization} &\textbf{ Few-shot Learning} & \textbf{Synthetic Tasks} &\textbf{Average} \\ \midrule
Llama2-7B-Chat~\cite{touvron2023llama} & 25.54 & 18.47 & 23.37 & 51.78 & 3.94 & 24.62 \\
    \ + PI~\cite{chen2023extending} & 23.42 & 23.73 & 25.34 & 51.63 & 7.63 & 26.35 \\ 
    \ + NTK~\cite{ntk} & 24.73 & 23.67 & 25.41 & 51.97 & 8.33 & 26.82 \\ 
    \ + Ms-PoE~\cite{zhang2024found} & 23.68 & \textbf{24.59} & 25.33 & 51.66 & 8.04 & 26.66 \\ 
    \ + AB~\cite{chen2023fortify} & \textbf{27.06} & 22.94 & 25.52 & \textbf{52.84} & 8.62 & 27.40 \\ 
    \ + MoICE (Ours) & 26.31 & 23.70 & \textbf{25.60} & 52.34 & \textbf{9.71} & \textbf{27.53} \\ \midrule \midrule
    Mistral-7B-Instruct-8k~\cite{jiang2023mistral} & 27.20 & 19.89 & 24.22 & 52.41 & 5.06 & 25.76 \\ \midrule
    \ + PI~\cite{chen2023extending} & 30.94 & \textbf{24.94} & 26.24 & 49.34 & \textbf{9.35} & 28.16 \\ 
    \ + NTK~\cite{ntk} & 30.46 & 21.21 & 23.89 & 52.41 & 8.44 & 27.28 \\ 
    \ + Ms-PoE~\cite{zhang2024found} & 27.90 & 17.89 & 20.28 & 48.59 & 8.95 & 24.72 \\ 
    \ + AB~\cite{chen2023fortify} & 29.81 & 21.95 & 25.58 & 54.42 & 7.89 & 27.93 \\ 
    \ + MoICE (Ours) & \textbf{31.09} & 22.98 & \textbf{26.69} & \textbf{55.76} & 8.02 & \textbf{28.91} \\ \midrule \midrule
    Qwen1.5-7B-Chat~\cite{qwen} & 34.66 & 35.91 & 25.77 & 56.89 & 33.83 & 37.41 \\ \midrule
    \ + PI~\cite{chen2023extending} & 28.28 & 17.08 & 24.60 & \textbf{57.51} & 32.67 & 32.03 \\
    \ + NTK~\cite{ntk} & 31.35 & 23.98 & 24.95 & 56.64 & 32.50 & 33.88 \\
    \ + Ms-PoE~\cite{zhang2024found} & OOM & OOM & OOM & OOM & OOM & N/A \\ 
    \ + AB~\cite{chen2023fortify} & OOM & OOM & OOM & OOM & OOM & N/A \\ 
    \ + MoICE (Ours) & \textbf{39.37} &\textbf{37.35} & \textbf{25.81} & 57.29 & \textbf{34.83} & \textbf{38.93} \\
    
 \bottomrule   
    
\end{tabular}
\end{threeparttable}}
\end{table}

MoICE consistently shows improved average performance across models with 4k, 8k, and 32k context lengths, surpassing previous competitive approaches.

\section{Attention waveforms}
\label{apx:attn-wave}

In this section, we will elaborate on attention waveforms and the concept of complementarity. 
Assuming $\mathbf{\hat{q}}_n^{h} \cdot \mathbf{\hat{k}}_m^{h}$ is the attention score (before softmax) of the $n$-th position to $m$-th position on the $h$-th attention head. The attention score can be formulated as follows: 

\begin{align*}
\mathbf{\hat{q}}_n^{h} \cdot \mathbf{\hat{k}}_m^{h}&=\left(\mathbf{R}_{\Theta,n}\mathbf{q}_n\right)^T\left(\mathbf{R}_{\Theta,m}\mathbf{k}_m\right) \\ 
&=  \operatorname{Re}\left[\sum_{j=0}^{d / 2-1} \mathbf{q_{n}^{h}}[2 j: 2 j+1] \mathbf{k_{m}^{h*}}[2 j: 2 j+1] e^{i(n-m) \theta_j}\right] \\
&=\sum_{j=0}^{d / 2-1}\left({q_n^{h}}_{2 j} \cdot {k_m^{h}}_{2 j}+{q_n^{h}}_{2 j+1} \cdot {k_m^{h}}_{2 j+1}\right) \cos \left((n-m) \theta_j\right) \\
& +\left({q_n^{h}}_{2 j} \cdot {k_m}_{2 j+1}-{q_n^{h}}_{2 j+1} \cdot {k_m^{h}}_{2 j}\right) \sin \left((n-m) \theta_j\right),
\end{align*}

where $\theta_j = B^{-2j/d}$, $B$ is the rotary base of RoPE. 
Considering a context-awareness task, basic context awareness relies on attending to the same token and then copying its next token as outputs~\cite{olsson2022context}. 
To simplify the calculation, we set both $\mathbf{q}_n^{h}$ and $\mathbf{k}_m^{h}$ as all-one vectors to observe the impact of relative positions on the attention when retrieving the same token from the context.
This impact (or the intensity of attention) is dubbed as attention waveform $\mathcal{W}$ by~\cite{chen2023fortify}.

\begin{equation*}
    \mathcal{W} \leq \sum_{j=0}^{d / 2-1} 2 \cos \left((n-m)\theta_j\right) .
\end{equation*}

As illustrated in Figure ~\ref{fig:waveform}, the waveform exhibits two notable mathematical properties concerning attention scores: it demonstrates fluctuations and undergoes a gradual decay with the increasing relative position (i.e., long-term decay). 

Chen et al.~\cite{chen2023fortify} observed that crucial information falling within the troughs of a waveform might diminish the performance of models employing RoPE. 
Meanwhile, they pointed out the waveform, characterized by peaks and troughs, vary across RoPE bases.
When leveraging the peaks of one attention wave to compensate for the overlook of the troughs in another, the model's capability to perceive and process information from diverse contextual positions can be enhanced. 
When a set of bases possesses this waveform characteristic, they are termed ``complementary.''

\section{Experimental details on pretraining}
\label{apx:pretraining}
In this section, we provide detailed experimental setup in Section~\ref{sec:pretraining}. The model we has 4 layers, 6 heads per layer, a hidden layer dimension of 512 and an intermediate size of 1280. We train the model using the OpenWebText pretraining~\cite{Gokaslan2019OpenWeb} dataset. We use four GTX A800-80Gs for training for 600k steps, with a context window of 512. During pretraining, we use a learning rate of 0.004 with a cosine annealing schedule and 6,000 warm-up steps.

\section{Discussions on more position embeddings}
\label{apx:other-pos}
In this section, we discuss other position embeddings and demonstrate why they are not studied, e.g., discarded in LLMs, do not exhibit attention waveform pattern, or are in the same family of RoPE:
Firstly, the waveform pattern only exists in position embeddings constructed by cosine functions. Regarding the cosine embedding used in the original Transformer, it does exhibit long-term decay and periodic waveforms. 
However, this embedding is disregarded in modern LLMs. 
Moreover, these embeddings are incorporated before the initial model layer rather than during the attention computation, making it hard to assess their impact on attention patterns.
Secondly, the learned positional embeddings utilized in BERT~\cite{devlin-etal-2019-bert} lack mathematical constraints to display periodic patterns. 
They are similarly added before the first model layer.
Thirdly, Alibi~\cite{press2022train} introduces a linear bias to attention scores. 
The linear bias is devoid of wave patterns.
The remaining popular positional embeddings used in LLMs such as xPos~\cite{sun-etal-2023-length} are RoPE-based variants. 
These variants are predominantly modified for long-context extrapolation rather than better context awareness. 
Therefore, they share the same shortcoming: tokens in attention trough are less focused on, thereby limiting context awareness, which is the study focus in our paper.

\section{Details on expert sets}
\label{apx:base-set}
Utilizing the RoPE-base searching algorithm as proposed by Chen et al.~\cite{chen2023fortify}, Table~\ref{tab:base-set} illustrates the resulting sets for different values of $N$.
\begin{table}[ht]
    \centering
    \caption{Searched Sets for Different $N$}
    \begin{tabular}{c|c}
    \toprule
        $N$ & \textbf{Searched Set} \\\midrule
        3 & \{10,000, 18,000, 19,000\} \\
        5 & \{10,000, 17,500, 18,000, 19,000, 20,000\}\\
        7 & \{10,000, 17,500, 18,000, 19,000, 20,000, 22,500, 25,000\} \\
        9 & \{10,000, 13,500, 17,500, 18,000, 19,000, 20,000, 22,500, 24,000, 25,000\}\\\bottomrule
    \end{tabular}
    \label{tab:base-set}
\end{table}

%%%%%%%%%%%%%%%%%%%%comment out when arxiv%%%%%%%%%%%%%%%%%%%%%%%%%%%%%%
\section{Limitations}
In this paper, we introduce a plug-in module called MoICE, which is integrated into the attention heads of open-source LLMs to enhance their context awareness. One limitation is that, due to limited computational resources, we did not investigate the effectiveness of pretraining a language model with more parameters using the MoICE architecture. 
Furthermore, our proposed method exploits the potential for context awareness within LLMs, but it does not imbue the models with additional inherent context awareness abilities. Achieving this may necessitate more extensive data to train all model parameters.

\section{Broader impacts and safety issues}
Our novel lightweight plug-in approach efficiently enhances the context awareness of open-source LLMs. 
This advancement holds great promise for enhancing the effectiveness of LLMs across diverse scenarios characterized by extensive and complex contexts, such as RAG, tool utilization, and role-playing.
The safety issue of our method mainly comes from the large language models we used, as they might output toxic and biased texts, which is a common safety issue regarding LLM research.

\begin{figure}[t]
    \centering
    \includegraphics[width=\linewidth]{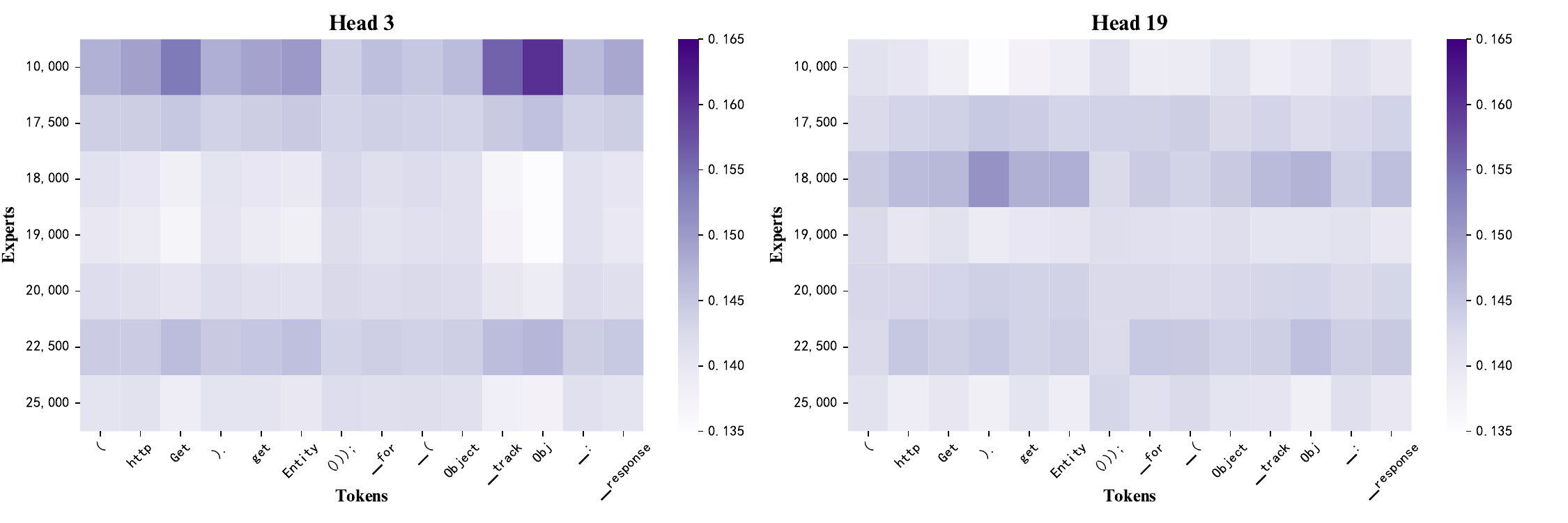}
    \caption{The routing weights across two distinct attention heads at the 27th layer in Llama-2-7B-chat. 
    The input tokens are randomly sampled from the training data, and the attention heads under observation are also randomly selected. 
    The horizontal axis depicts the input tokens, while the vertical axis represents experts with varying RoPE angles.
    Due to their distinct functions, each head dynamically chooses different experts to process individual tokens.
    Input text can be found in Figure ~\ref{fig:router_input}.}
    \label{fig:router}
\end{figure}

\begin{figure}[t]
    \centering
    \includegraphics[width=0.6\linewidth,height=0.5\linewidth]{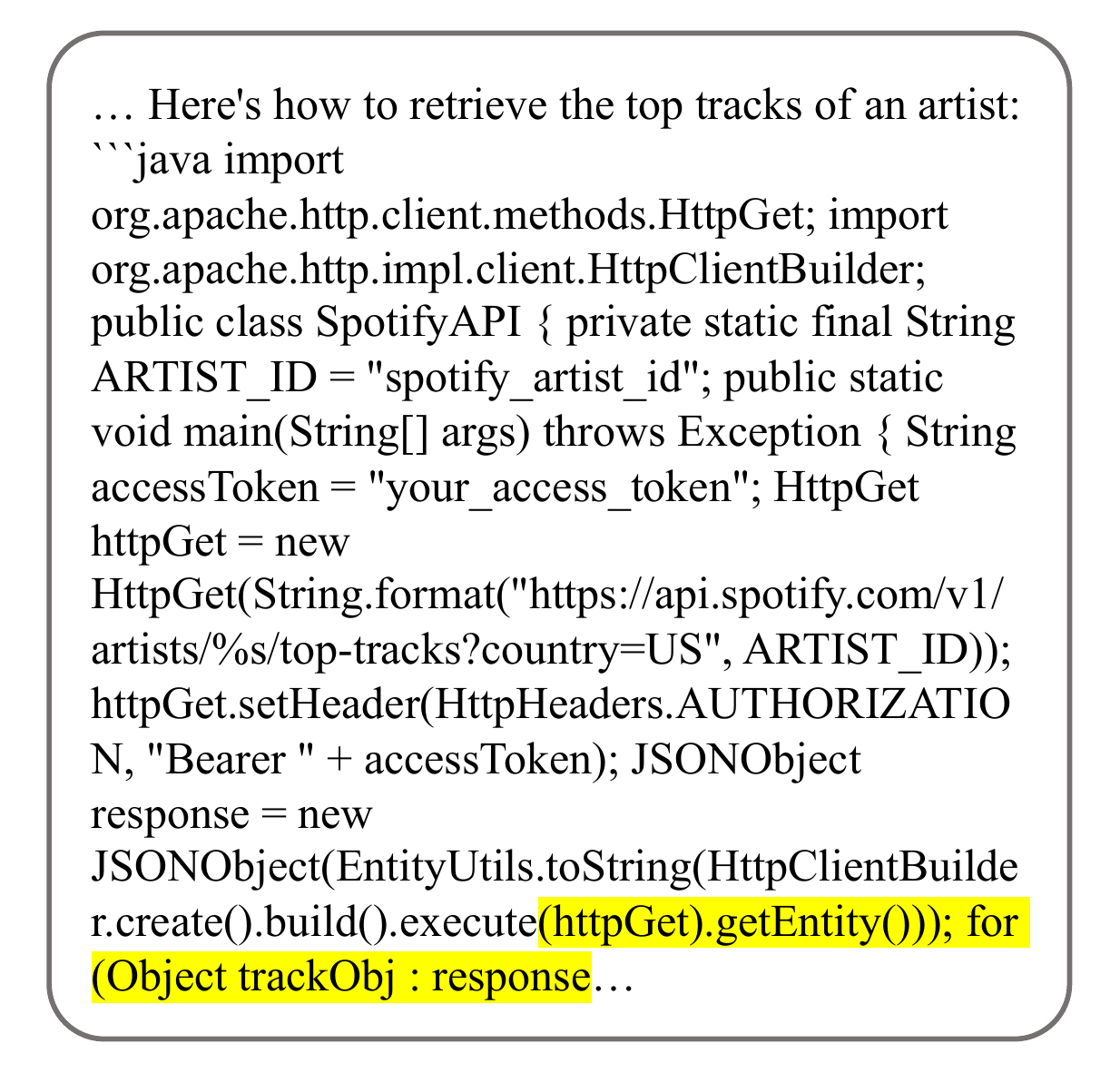}
    \caption{The input text in Figure ~\ref{fig:router}. To clearly display, we only show part of the input text, where the text with a yellow background corresponds to the decoded tokens.}
    \label{fig:router_input}
\end{figure}

\begin{figure}[t]
    \centering
    \includegraphics[width=0.7\linewidth]{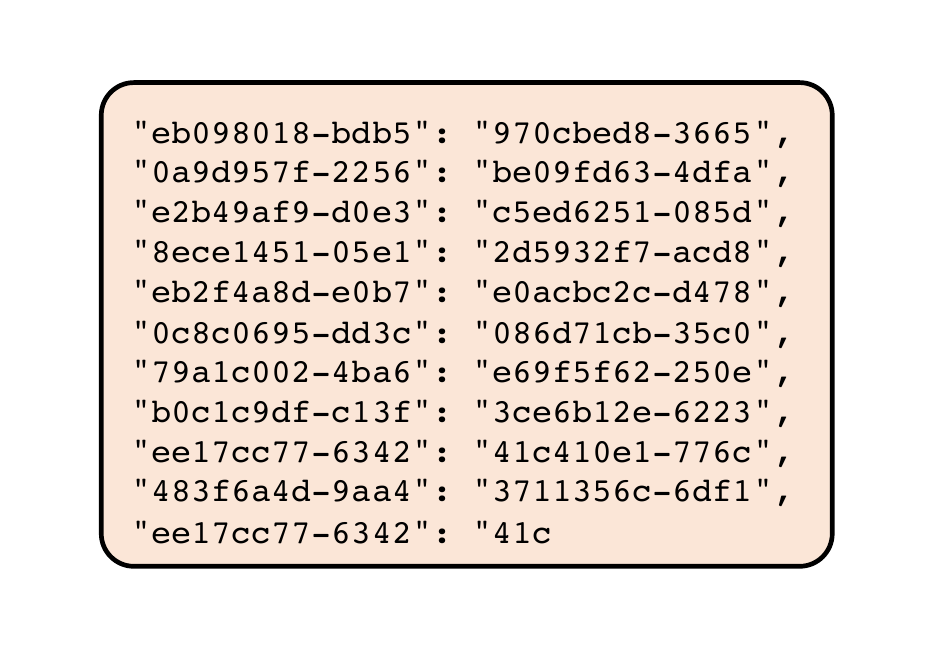}
    \caption{The input prompt example in Section~\ref{sec:pretraining}. We use 10 key-value pairs as examples in prompt, which includes a query key. We insert the query key-value pair in different positions of examples (In the prompt example above, the query key is inserted in the 9th position). The model's task is to find the value corresponding to the query key and output it, which evaluates its ability of context awareness. }
    \label{fig:kv_prompt}
\end{figure}

\begin{table}[ht]
\centering
\caption{The mean and standard deviation of MoICE. We repeat L-eval~\cite{an2023eval} experiments 5 times with different random seeds. The randomness of MoICE results from the initialization of MoICE router when training, which causes slight differences in performance.}
\label{table:mean}
\resizebox{\linewidth}{!}{
\begin{threeparttable}
\begin{tabular}{l|c|c|c|c|c|c|c|c}
\toprule
\multirow{2}*{\textbf{Method}} & \multicolumn{5}{c}{\textbf{Closed - Ended Task}} & \multicolumn{3}{c}{\textbf{Open - Ended Task}}\\ \cmidrule(r){2-6} \cmidrule(r){7-9}
                     ~ & \multicolumn{1}{c}{\textbf{Coursera}} & \multicolumn{1}{c}{\textbf{QuALITY}} & \multicolumn{1}{c}{\textbf{TOEFL}} & \multicolumn{1}{c}{\textbf{SFiction}} & \multicolumn{1}{c}{\textbf{Average}} & \multicolumn{1}{c}{\textbf{wins}} & \multicolumn{1}{c}{\textbf{ties}} & \multicolumn{1}{c}{\textbf{win-rate\%}} \\ \midrule

Llama2-7B-Chat~\cite{touvron2023llama} & 36.77 $\pm$ 0.00 & 38.12 $\pm$ 0.00 & 55.02 $\pm$ 0.00 & 60.16 $\pm$ 0.00 & 47.52 $\pm$ 0.00 & 68.00 $\pm$ 0.00 & 117.00 $\pm$ 0.00 & 34.94 $\pm$ 0.00 \\ \midrule
+ MoICE & 39.65 $\pm$ 0.32 & 41.88 $\pm$ 0.27 & 56.28 $\pm$ 0.21 & 64.84 $\pm$ 0.00 & 50.66 $\pm$ 0.05 & 89.00 $\pm$ 1.00 & 117.20 $\pm$ 1.48 & 40.77 $\pm$ 0.20 \\ \midrule \midrule
Mistral-7B-Instruct-8k~\cite{jiang2023mistral} & 45.20 $\pm$ 0.00 & 44.06 $\pm$ 0.00 & 62.08 $\pm$ 0.00 & 61.72 $\pm$ 0.00 & 53.27 $\pm$ 0.00 & 71.00 $\pm$ 0.00 & 105.00 $\pm$ 0.00 & 34.11 $\pm$ 0.00 \\ \midrule
        + MoICE & 48.08 $\pm$ 0.24 & 46.73 $\pm$ 0.27 & 65.35 $\pm$ 0.81 & 62.18 $\pm$ 1.19 & 55.59 $\pm$ 0.16 & 85.00 $\pm$ 1.10 & 115.20 $\pm$ 2.05 & 39.39 $\pm$ 0.21 \\

 \bottomrule   
    
\end{tabular}

\end{threeparttable}}
\end{table}

\begin{table}[h]
\centering
\caption{The ablation study on the auxiliary loss of MoICE. To assess the impact of this loss term, we perform an ablation experiment on two LLMs by removing it from Eq.~\ref{eq:all_loss}. The results show a significant drop in performance, highlighting the positive impact of the auxiliary loss.}
\label{tab:aux_loss}
\resizebox{0.85\linewidth}{!}{
\begin{tabular}{@{}lccccc@{}}
\toprule
\textbf{Method} & \textbf{Coursera } & \textbf{QuALITY } & \textbf{TOEFL } & \textbf{SFiction } &  \textbf{Average } \\ \midrule
\textbf{Llama2-7B-chat}~\cite{touvron2023llama} 
\vspace{2pt} \\ 

\ \ \ \ \ MoICE w/o aux loss & 39.83 & 41.58 & 56.13 & 62.50 & 50.01 \\ 
\ \ \ \ \  MoICE w/ aux loss & 39.83 & 42.08 & 56.13 & 64.84 & 50.72 \\ \midrule \midrule
\textbf{Mistral-7B-Instruct-8k}~\cite{jiang2023mistral} \vspace{2pt} \\ 
\ \ \ \ \  MoICE w/o aux loss & 47.67 & 46.04 & 64.68 & 58.59 & 54.25 \\ 
\ \ \ \ \  MoICE w/ aux loss & 47.82 & 46.53 & 64.68 & 62.50 & 55.38 \\ \bottomrule
\end{tabular}}
\end{table}

\clearpage
\newpage
\section*{NeurIPS Paper Checklist}

\begin{enumerate}

\item {\bf Claims}
    \item[] Question: Do the main claims made in the abstract and introduction accurately reflect the paper's contributions and scope?
    \item[] Answer: \answerYes{} % Replace by \answerYes{}, \answerNo{}, or \answerNA{}.
    \item[] Justification: We propose an effective and efficient approach for enhancing the context awareness of LLMs.

\item {\bf Limitations}
    \item[] Question: Does the paper discuss the limitations of the work performed by the authors?
    \item[] Answer: \answerYes{} % Replace by \answerYes{}, \answerNo{}, or \answerNA{}.
    \item[] Justification: We create a separate "Limitations" section in Appendix.

\item {\bf Theory Assumptions and Proofs}
    \item[] Question: For each theoretical result, does the paper provide the full set of assumptions and a complete (and correct) proof?
    \item[] Answer: \answerNA{} % Replace by \answerYes{}, \answerNo{}, or \answerNA{}.
    \item[] Justification: Our paper does not introduce theoretical results.

    \item {\bf Experimental Result Reproducibility}
    \item[] Question: Does the paper fully disclose all the information needed to reproduce the main experimental results of the paper to the extent that it affects the main claims and/or conclusions of the paper (regardless of whether the code and data are provided or not)?
    \item[] Answer: \answerYes{} % Replace by \answerYes{}, \answerNo{}, or \answerNA{}.
    \item[] Justification: We’ve shared the link to the code. We promise to open source.

\item {\bf Open access to data and code}
    \item[] Question: Does the paper provide open access to the data and code, with sufficient instructions to faithfully reproduce the main experimental results, as described in supplemental material?
    \item[] Answer: \answerYes{} % Replace by \answerYes{}, \answerNo{}, or \answerNA{}.
    \item[] Justification: We promise to open code. We have posted an anonymous code link.

\item {\bf Experimental Setting/Details}
    \item[] Question: Does the paper specify all the training and test details (e.g., data splits, hyperparameters, how they were chosen, type of optimizer, etc.) necessary to understand the results?
    \item[] Answer: \answerYes{} % Replace by \answerYes{}, \answerNo{}, or \answerNA{}.
    \item[] Justification: See Section 4.

\item {\bf Experiment Statistical Significance}
    \item[] Question: Does the paper report error bars suitably and correctly defined or other appropriate information about the statistical significance of the experiments?
    \item[] Answer: \answerYes{} % Replace by \answerYes{}, \answerNo{}, or \answerNA{}.
    \item[] Justification: We report the p-value in the t-test in the ``Results and Analysis'' paragraph of Section 4.2.

\item {\bf Experiments Compute Resources}
    \item[] Question: For each experiment, does the paper provide sufficient information on the computer resources (type of compute workers, memory, time of execution) needed to reproduce the experiments?
    \item[] Answer: \answerYes{} % Replace by \answerYes{}, \answerNo{}, or \answerNA{}.
    \item[] Justification: See Section 4.
    
\item {\bf Code Of Ethics}
    \item[] Question: Does the research conducted in the paper conform, in every respect, with the NeurIPS Code of Ethics \url{https://neurips.cc/public/EthicsGuidelines}?
    \item[] Answer: \answerYes{} % Replace by \answerYes{}, \answerNo{}, or \answerNA{}.
    \item[] Justification: We have conformed the NeurIPS Code of Ethics.

\item {\bf Broader Impacts}
    \item[] Question: Does the paper discuss both potential positive societal impacts and negative societal impacts of the work performed?
    \item[] Answer: \answerYes{} % Replace by \answerYes{}, \answerNo{}, or \answerNA{}.
    \item[] Justification: See Section "Broader Impacts and Safeguards."
    
\item {\bf Safeguards}
    \item[] Question: Does the paper describe safeguards that have been put in place for responsible release of data or models that have a high risk for misuse (e.g., pretrained language models, image generators, or scraped datasets)?
    \item[] Answer: \answerYes{} % Replace by \answerYes{}, \answerNo{}, or \answerNA{}.
    \item[] Justification: See Section "Broader Impacts and Safety Issues."

\item {\bf Licenses for existing assets}
    \item[] Question: Are the creators or original owners of assets (e.g., code, data, models), used in the paper, properly credited and are the license and terms of use explicitly mentioned and properly respected?
    \item[] Answer: \answerYes{} % Replace by \answerYes{}, \answerNo{}, or \answerNA{}.
    \item[] Justification: We have correctly cited all the data, scripts, and models we used.

\item {\bf New Assets}
    \item[] Question: Are new assets introduced in the paper well documented and is the documentation provided alongside the assets?
    \item[] Answer: \answerYes{} % Replace by \answerYes{}, \answerNo{}, or \answerNA{}.
    \item[] Justification: We have a README document for our code.

\item {\bf Crowdsourcing and Research with Human Subjects}
    \item[] Question: For crowdsourcing experiments and research with human subjects, does the paper include the full text of instructions given to participants and screenshots, if applicable, as well as details about compensation (if any)? 
    \item[] Answer: \answerNA{} % Replace by \answerYes{}, \answerNo{}, or \answerNA{}.
    \item[] Justification: This paper does not involve crowdsourcing experiments.

\item {\bf Institutional Review Board (IRB) Approvals or Equivalent for Research with Human Subjects}
    \item[] Question: Does the paper describe potential risks incurred by study participants, whether such risks were disclosed to the subjects, and whether Institutional Review Board (IRB) approvals (or an equivalent approval/review based on the requirements of your country or institution) were obtained?
    \item[] Answer: \answerNA{} % Replace by \answerYes{}, \answerNo{}, or \answerNA{}.
    \item[] Justification: This paper does not involve crowdsourcing experiments.

\end{enumerate}

\end{document}